\documentclass{ifacconf}

\usepackage{graphicx}      % include this line if your document contains figures
\usepackage{natbib}        % required for bibliography
\usepackage{amsmath}       % 确保引入 amsmath 宏包
\usepackage{subfig}
\usepackage{booktabs}
\usepackage{array}
\usepackage{multirow} 
\usepackage{makecell}
\usepackage{xcolor}

\usepackage{gensymb}
\begin{document}
\begin{frontmatter}

\title{An Extended Generalized Prandtl-Ishlinskii Hysteresis Model for I$^2$RIS Robot } 
  

\thanks[footnoteinfo]{This work was supported by the U.S. National Institutes of Health under grant numbers R01EB023943 and R01EB034397 and partially by JHU internal funds.}

\author[First]{Yiyao Yue} 
\author[First]{Mojtaba Esfandiari} 
\author[First]{Pengyuan Du}
\author[Second]{Peter Gehlbach}
\author[Third]{Makoto Jinno}
\author[First]{Adnan Munawar}
\author[First]{Peter Kazanzides}
\author[First]{Iulian Iordachita}

\address[First]{Laboratory for Computational Sensing and Robotics,
    Johns Hopkins University, Baltimore, MD 21218 USA (e-mail: yyue19@jh.edu)}
    
\address[Second]{Wilmer Eye Institute, Johns Hopkins Hospital, Baltimore, MD, 21287 USA (e-mail: pgelbach@jhmi.edu) }

\address[Third]{School of Science and Engineering, Kokushikan University, Tokyo, 154-8515, Japan (e-mail: mjinno@kokushikan.ac.jp)}

\begin{abstract}                % Abstract of 50--100 words
Retinal surgery requires extreme precision due to constrained anatomical spaces in the human retina. To assist surgeons achieve this level of accuracy, the Improved Integrated Robotic 
Intraocular Snake (I\textsuperscript{2}RIS) with dexterous capability has been developed. However, such flexible tendon-driven robots often suffer from hysteresis problems, which significantly challenges precise control and positioning. In particular, we observed multi-stage hysteresis phenomena in the small-scale I\textsuperscript{2}RIS. In this paper, we propose an Extended Generalized Prandtl-Ishlinskii (EGPI) model to increase the fitting accuracy of the hysteresis. The model incorporates a novel switching mechanism that enables it to describe multi-stage hysteresis in the regions of monotonic input. Experimental validation on I\textsuperscript{2}RIS data demonstrates that the EGPI model outperforms the conventional Generalized Prandtl–Ishlinskii (GPI) model in terms of RMSE, NRMSE, and MAE across multiple motor input directions. The EGPI model in our study highlights the potential in modeling multi-stage hysteresis in minimally invasive flexible robots.
\end{abstract}

\begin{keyword}
Robotic retinal microsurgery, medical robotics, flexible intraocular robot, hysteresis modeling, Prandtl–Ishlinskii model.
\end{keyword}

\end{frontmatter}
%===============================================================================

%%%%%%%%%%%%%%%%%%%%%%%%%%%%%%%%%%%%%%%%%%%%%%%%%%%%%%%%%%%%%%%%%%%%%%%%%
%%%%%%%%%%%%%%%%%%%%%%%%%%%%%%%%%%% INTRODUCTION %%%%%%%%%%%%%%%%%%%%%%%%
%%%%%%%%%%%%%%%%%%%%%%%%%%%%%%%%%%%%%%%%%%%%%%%%%%%%%%%%%%%%%%%%%%%%%%%%%

\section{Introduction}
\begin{figure}[!t]
    \centering
    \centerline{\includegraphics[width= 1.0 \columnwidth]{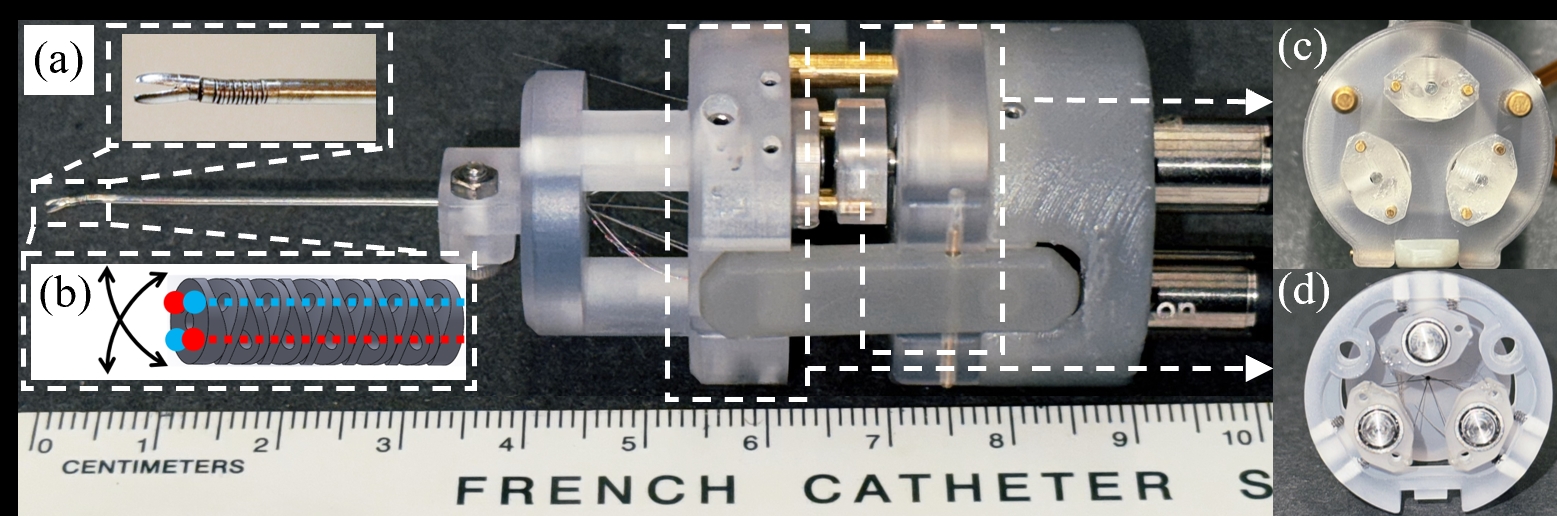}}
    \caption{Overview of the I$^2$RIS robot. (a) The I$^2$RIS robot and its dexterous distal unit. (b) Conceptual design of dexterous distal unit. (c) Cross-section of the actuation unit. (d) Cross-section of the instrument unit.}
    \label{fig:robot_design}
\end{figure}

% 角膜剥离手术-病症现状-相对应的机器人系统-机器人系统对应的建模难题-连续体机器人迟滞与补偿的的研究-本文贡献-rest of the article结构
% Treating the causal factors of retinal diseases requires performing delicate microsurgical procedures. For example, to treat such retinal diseases as diabetic macular edema (DME) (\cite{hu2018efficacy}) and epiretinal membrane (ERM) (\cite{stevenson2016epiretinal}), surgeons try to carefully peel off the internal limiting membrane (ILM) or ERM using microforces (\cite{charles2003techniques}). Of note, "DME is a leading cause of vision loss in persons with diabetes mellitus (DM)" (\cite{varma2014prevalence}), and about 347 million people worldwide suffer from DM (\cite{danaei2011national}). In addition, about 30 million people in the US, at least in one eye, have an ERM (\cite{klein1994epidemiology}).
Delicate microsurgical procedures are required to treat retinal diseases such as diabetic macular edema (DME) and epiretinal membrane (ERM), where surgeons must carefully peel the internal limiting membrane (ILM) or ERM using microforceps (\cite{charles2003techniques}). DME is a major cause of vision loss among diabetic patients (\cite{varma2014prevalence}). ERM is also prevalent, impacting approximately 30 million people in the U.S. alone (\cite{klein1994epidemiology}). The human retina’s small thickness makes such procedures highly challenging. Surgeons' physiological hand tremor (182$\,\mu$m, \cite{riviere2000study}) and patients' involuntary eye motion caused by respiration and heartbeat (81$\,\mu$m, in amplitude and 1$\,$Hz, in frequency \cite{de2011heartbeat}) hinder precise control, increasing the risk of errors like accidental nerve fiber pinching and hemorrhages (\cite{lumi2022simple}). To address these challenges, robotic systems such as Intraocular Robotic Interventional and Surgical System (IRISS) (\cite{rahimy2013robot}), and the Steady-Hand Eye Robot (SHER) (\cite{uneri2010new}) have been developed. The SHER is a five-degrees-of-freedom (5-DoF) manipulator that provides tremor cancellation. It lacks the necessary dexterity to perform complex operations, such as ERM peeling. To overcome this limitation, the Improved Integrated Robotic Intraocular Snake (I\textsuperscript{2}RIS) was designed (\cite{jinno2021improved}), adding two additional DoFs-yaw and pitch bending, along with a gripping function. (see Fig.~\ref{fig:robot_design}).

Unlike conventional rigid robotic manipulators, continuum and snake-like robots present unique challenges in modeling and control (\cite{iqbal2025continuum}) due to their hysteresis caused by mechanical complexity and nonlinear structural behavior (\cite{esfandiari2024data}). Hysteresis significantly influences accurate modeling and control, making it one of the primary bottlenecks in precision applications. To address this issue, data-driven approaches such as neural networks and classical modeling techniques have been explored. For example, long short-term memory (LSTM) networks have been applied for hysteresis modeling in robotic catheters (\cite{wu2021hysteresis}) and continuum robots (\cite{wang2024using}). However, these methods often lack physical meaning and typically require large training datasets, which are not practical for I\textsuperscript{2}RIS. Other classical data-driven methods, such as the Bouc-Wen model (\cite{zhang2023hysteresis}), the GPI model (\cite{al2010analytical}), and the Modified GPI model (\cite{gao2023modeling}), have limited flexibility in capturing multi-stage hysteresis behavior.

In particular, our experimental data from the I\textsuperscript{2}RIS robot (see Fig.~\ref{fig_m1}(a)) shows the existence of a multi-stage hysteresis behavior, primarily caused by the cable-driven actuation mechanism. In addition, the small size of the robot, restricts the embedding of the sensing modules, causing the accurate position control of the robot to be more challenging (\cite{esfandiari2024data}. Hence, having an accurate model of the I$^2$RIS hysteresis is necessary to accurately control the robot and improve patient safety.

% \begin{figure}[!t]
%     \centering
%     \centerline{\includegraphics[width= 1.0 \columnwidth]{fig/SHER_I2RIS_Real.pdf}}
%     \caption{The I$^2$RIS attached to the SHER 3.0 handle.}
%     \label{fig:SHER_I2RIS_Real}
% \end{figure}

To tackle the multi-stage hysteresis modeling problem of the I$^2$RIS robot, we proposed an Extended Generalized Prandtl–Ishlinskii model. While the classical GPI model provides a strong foundation due to its capability to model asymmetric hysteresis and its analytically invertible structure, it falls short in representing the multi-stage transitions present in our system. The EGPI model builds upon the GPI framework by introducing a switching mechanism to capture hysteresis and its transitions more accurately. The contributions of this work are as follows:  

\begin{itemize}
    \item We designed an EGPI model to describe the input and output relationship that shows multi-stage hysteresis behavior in the region of monotonic changes in input. 
    \item We characterized the multi-stage hysteresis behavior of the I$^2$RIS robot by using the proposed EGPI model. This model considers the mapping between motor encoder values and the robot's pitch and yaw bending angles. 
    \item We optimized the parameters of the EGPI model using experimental data from the I$^2$RIS robot by the Levenberg-Marquardt(LM) method.  
    
\end{itemize}

% The effectiveness of the proposed method in modeling the hysteresis characteristics of the I$^2$RIS robot, which outperforms the conventional GPI model, is evaluated using experimental results. 

The effectiveness of the proposed method in modeling the hysteresis characteristics of the I$^2$RIS robot, is evaluated using experimental results.

%%%%%%%%%%%%%%%%%%%%%%%%%%%%%%%%%%%%%%%%%%%%%%%%%%%%%%%%%%%%%%%%%%%%%%%%%
%%%%%%%%%%%%%%%%%%%%%%%%%%%%%%%%%%% Method %%%%%%%%%%%%%%%%%%%%%%%%%%%%%%
%%%%%%%%%%%%%%%%%%%%%%%%%%%%%%%%%%%%%%%%%%%%%%%%%%%%%%%%%%%%%%%%%%%%%%%%%

\section{Materials and Methods}
\subsection{I\textsuperscript{2}RIS Design}
The I\textsuperscript{2}RIS (see Fig.~\ref{fig:robot_design}(a)) is a 2-DoF snake-like robot, capable of bending $\pm 45 \degree$ for pitch and yaw directions, and equipped with a micro-gripper mounted on the distal segment (0.9$\,$mm diameter and 3.0$\,$mm length) (\cite{esfandiari2024data}). Fig.~\ref{fig:robot_design}(b) shows the conceptual design of the dexterous distal unit of the I\textsuperscript{2}RIS. The dexterous distal unit is composed of 12 disk-like elements, providing 2-DoF bending joints actuated by four wires. The wire driven mechanism is actuated by the rotational motion of pulleys, whose cross-sections are shown in Fig.~\ref{fig:robot_design}(c) and Fig.~\ref{fig:robot_design}(d). Further details on the dexterous distal unit and the drive unit design can be found in \cite{jinno2021improved}.

\subsection{Extended Generalized Prandtl-Ishlinskii Model}
The GPI model (\cite{al2010analytical}) combines a generalized play operator with a density function to characterize hysteresis behavior. The play operator $G_r[v](t)$ (see Fig.~\ref{fig_m1}(b)) employs two envelope functions, $\gamma_l$ and $\gamma_r$, to represent the ascending and descending branches of the hysteresis loop, respectively. The parameter $r$ denotes the magnitude of the backlash, and each GPI play operator is defined for a specific value of $r$ with the form:
\begin{figure*}[!t]
\centering
\subfloat[]{\includegraphics[width=1.72in]{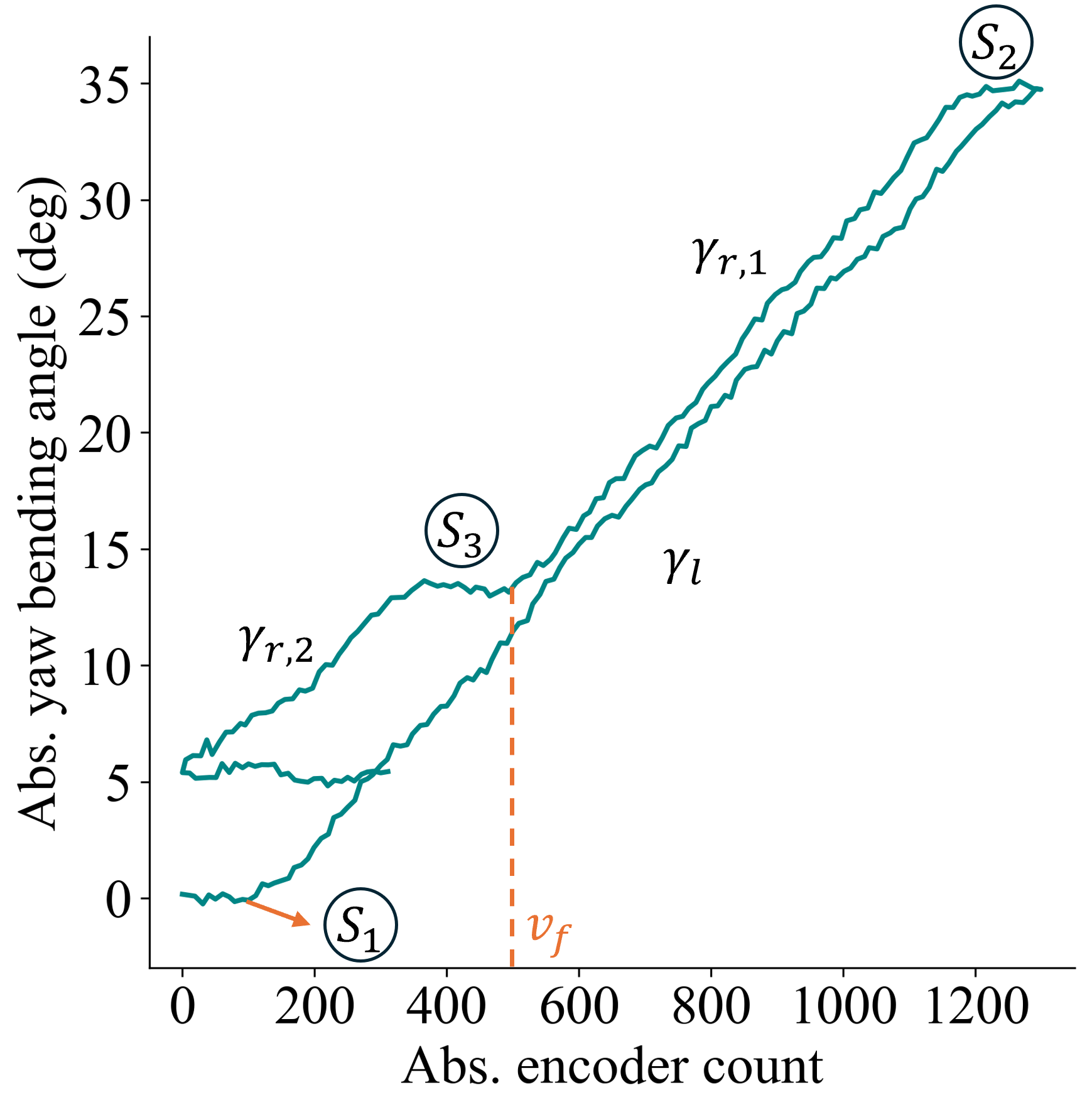}%
\label{fig_m1_a}}
\hfil
\subfloat[]{\includegraphics[width=1.72in]{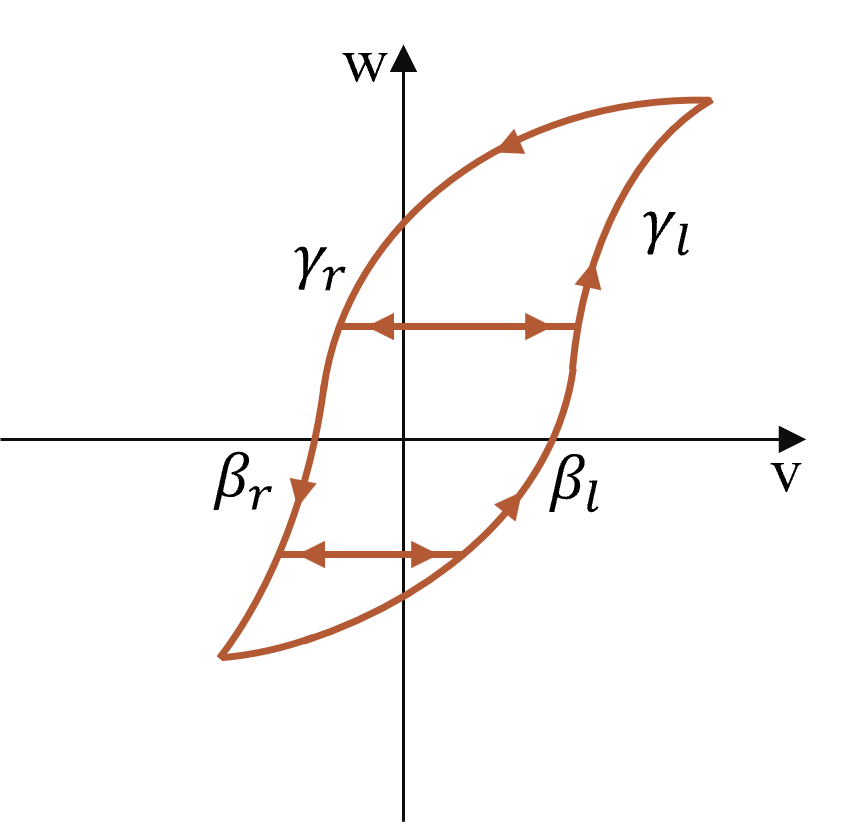}%
\label{fig_m1_b}}
\hfil
\subfloat[]{\includegraphics[width=1.72in]{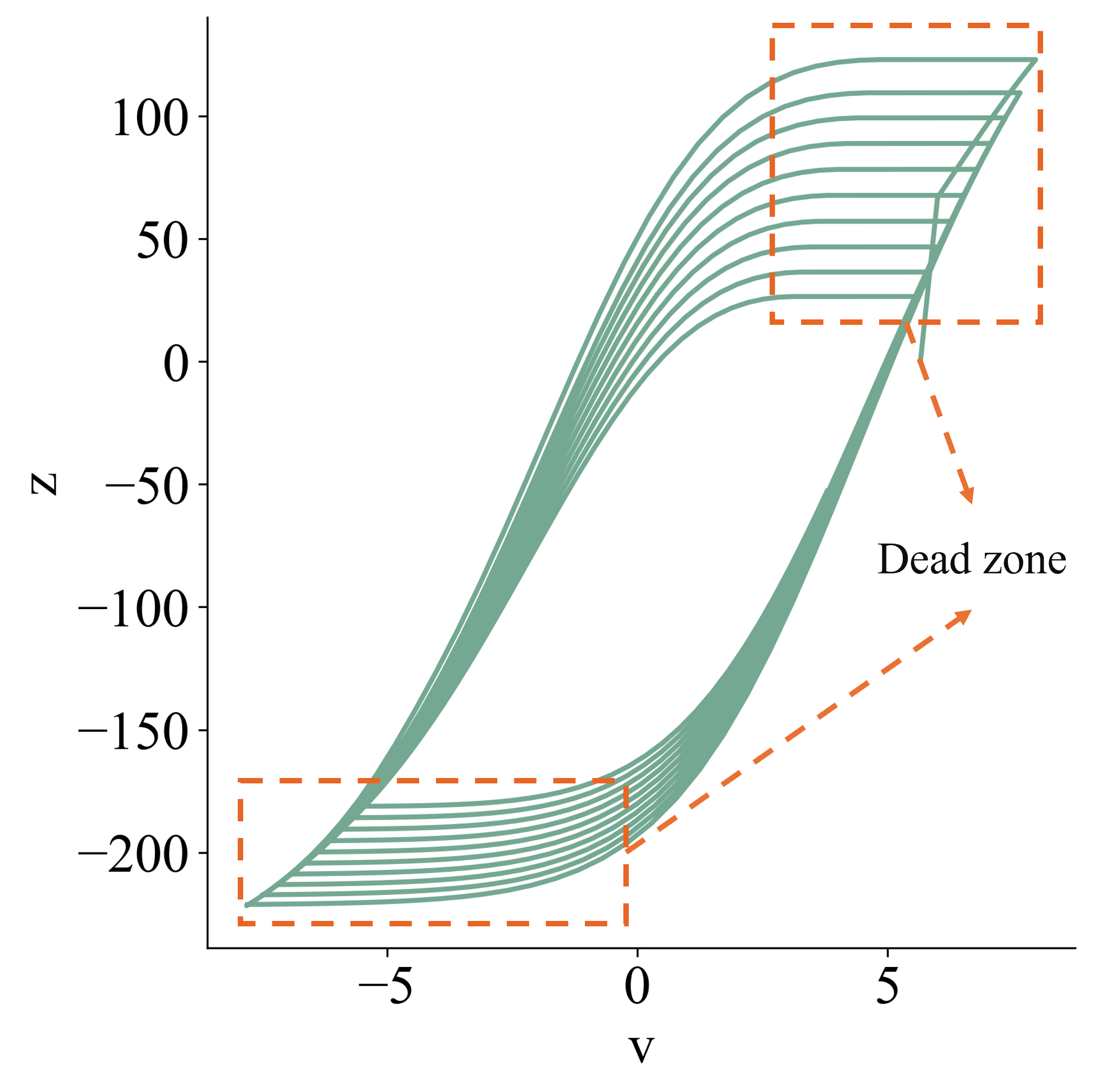}%
\label{fig_m1_c}}
\hfil
\subfloat[]{\includegraphics[width=1.72in]{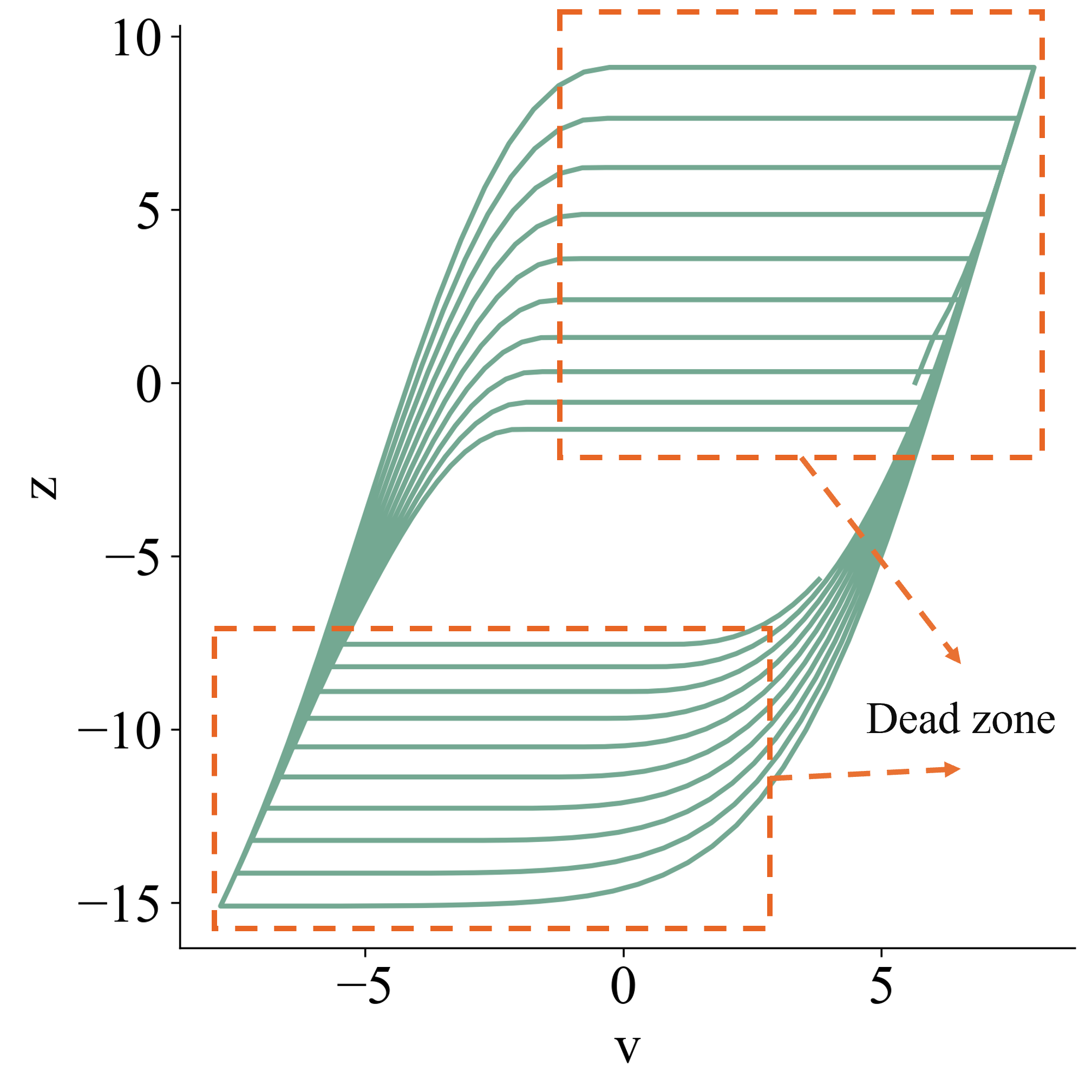}%
\label{fig_m1_d}}
\caption{(a) An example of I\textsuperscript{2}RIS data with multi-stage hysteresis $S_1$,$S_2$ and $S_3$: Absolute yaw bending angle as a function of absolute yaw motor encoder value. (b) An example of the GPI operator. (c) Output of the GPI model $z_1$. (d) Output of the GPI model $z_2$.}
\label{fig_m1}
\end{figure*}
\begin{equation} \label{eq:sample}
G_r[v](t) =
\begin{cases} 
\text{max}(\gamma_l (v(t))-r),w(t_i)), & \text{for } \dot{v}(t)>0 \\ 
\text{min}(\gamma_r (v(t))+r),w(t_i)), & \text{for }\dot{v}(t)<0 \\ 
w(t_i), & \text{for } \dot{v}(t)=0.
\end{cases}
\end{equation}
The zero points of the GPI play operator $\beta_l$ and $\beta_r$ are defined by two envelope functions (\cite{al2010analytical}):
\begin{equation} \label{eq:sample}
\begin{cases} 
\beta_l=\gamma_l^{-1}(r), & \text{for } \dot{v}(t)>0 \\
\beta_r=\gamma_r^{-1}(-r), & \text{for } \dot{v}(t)<0.
\end{cases}
\end{equation}
The difference in magnitude between these two zero points enables the GPI play operator to model asymmetrical hysteresis loops(\cite{al2010analytical}). Moreover, the envelope functions $\gamma_l$ and $\gamma_r$ satisfy the Lipschitz continuity condition (\cite{brokate2012hysteresis}), ensuring the monotonicity of the increasing and decreasing branches of the play operator.
The basic idea of the final output $y(t)$ of the GPI model is obtained by taking a finite weighted sum of play operators with different backlash parameters $r$ (\cite{al2010analytical}). Accordingly, the input-output relationship between $[v](t)$ and $y(t)$  can be expressed as:
\begin{equation} \label{eq:sample}
y(t)=\Pi[v](t)=\sum_{i=0}^{n} p(r_i)G_r[v](t)
\end{equation}
where $n$ denotes the number of different play operators, and the weighting factor $p(r_i)$ commonly referred to as the density function. An example of the GPI model output is shown in Fig.~\ref{fig_m1}(c).

However, as shown in Fig.~\ref{fig_m1}(a), the decreasing segment (left side) of the I\textsuperscript{2}RIS data exhibits a piecewise monotonically decreasing pattern, indicating the existence of multiple stages of backlash (\cite{lee2021non}) within a single monotonic phase of motion. Since the GPI model can only capture hysteresis loops characterized by a single stage of backlash, we propose an extension, called the Extended Generalized Prandtl-Ishlinskii (EGPI) model, to address this limitation. The EGPI model treats the presence of multiple hysteresis stages as a switching behavior between different GPI models. In other words, it can be interpreted as the system selecting among different GPI models. It is important to note that these play operators of different GPI models are not merely differentiated by the parameter $r$, but are fundamentally different, each associated with its own pair of envelope functions. Next, we derive the EGPI model with two backlash stages in the monotonically increase and decrease parts, respectively. Two different play operators $H_{r,1}[v](t)$ and $H_{r,2}[v](t)$ are expressed as:
\begin{equation} \label{eq:sample}
H_{r,1}[v](t) =
\begin{cases} 
\text{max}(\gamma_{l,1} (v(t))-r),w(t_i)), & \text{for } \dot{v}(t)>0 \\ 
\text{min}(\gamma_{r,1} (v(t))+r),w(t_i)), & \text{for } \dot{v}(t)<0 \\ 
w(t_i), & \text{for } \dot{v}(t)=0
\end{cases}
\end{equation}
\begin{equation} \label{eq:sample}
H_{r,2}[v](t) =
\begin{cases} 
\text{max}(\gamma_{l,2} (v(t)) - \kappa_1 r),w(t_i)), & \text{for } \dot{v}(t)>0 \\ 
\text{min}(\gamma_{r,2} (v(t)) + \kappa_2 r),w(t_i)), & \text{for } \dot{v}(t)<0 \\ 
w(t_i), & \text{for } \dot{v}(t)=0
\end{cases}
\end{equation}

where $\gamma_{l,1}$ and $\gamma_{r,1}$ in the first play operator are different from $\gamma_{l,2}$ and $\gamma_{r,2}$. The $\kappa_1$ and $\kappa_2$ are two magnitude regulators (\cite{gao2023modeling}) to adjust the backlash magnitude in the second stage. Two GPI model output $z_1(t)$ and $z_2(t)$ are expressed as:
\begin{equation} \label{eq:sample}
z_1(t)=\Phi_1[v](t)=\sum_{i=0}^{n} p(r_i)H_{r,1}[v](t)
\end{equation}
\begin{equation} \label{eq:sample}
z_2(t)=\Phi_2[v](t)=\sum_{i=0}^{n} p(r_i)H_{r,2}[v](t).
\end{equation}

The key idea of this work lies in the output strategy of the EGPI model $z(t)$, which enables switching between different GPI models. Specifically, the model transitions from $\Phi_1[v](t)$ to $\Phi_2[v](t)$ at designated flag points: $v_{f,1}$ during the increasing part and $v_{f,2}$ during the decreasing phase (see Fig.~\ref{fig:EGPI_sim}). When modeling the I\textsuperscript{2}RIS hysteresis behavior, the flag point is determined from the experimental data, which will be introduced in Section 3.
\begin{equation} \label{eq:sample}
z(t) = \Phi[v](t)
     = \begin{cases} 
\Phi_1[v](t), & \text{for } \dot{v}(t)>0 \land v(t)<v_{f,1}\\ 
\Phi_2[v](t), & \text{for } \dot{v}(t)>0 \land v(t)\geq v_{f,1} \\ 
\Phi_1[v](t), & \text{for } \dot{v}(t)\leq0 \land v(t)> v_{f,2} \\ 
\Phi_2[v](t), & \text{for } \dot{v}(t)\leq0 \land v(t)\leq v_{f,2}
\end{cases}
\end{equation}

This novel switching mechanism allows the EGPI model to capture more than one hysteresis stage that the single GPI model fails to represent.

%%%%%%%%%%%%%%%%%%%%%%%%%%%%%%%%%%%%%%%%%%%%%%

% \section{Experimental Setup}

% \section{Results}

\subsection{Simulation Verification}
For simulation verification, we manually selected model parameters (\cite{al2010analytical}) to qualitatively illustrate the performance of the EGPI model. An input signal $v(t)$ is used to test the output $z(t)$ of the EGPI model, along with the outputs of two underlying GPI models, $z_1(t)$, $z_2(t)$, which together form $z(t)$. The input signal $v(t)$ is taking the decaying sinusoids wave (\cite{wu2021hysteresis}) of the form:
\begin{equation} \label{eq:sample}
v(t) = 8e^{-0.04t}(\sin(2\pi t+ \frac{\pi}{4}))
\end{equation}
where $t\in[0,10]$. The envelope functions of two different play operators $H_{r,1}[v](t)$ and $H_{r,2}[v](t)$ are set as:
\begin{align}
\gamma_{l,1}(v) &= 8\tanh(0.2v - 0.5) \\
\gamma_{r,1}(v) &= 9\tanh(0.2v - 0.1) \\
\gamma_{l,2}(v) &= 8\tanh(0.2v - 1) \\
\gamma_{r,2}(v) &= 10\tanh(0.2v + 0.5) + 0.1.
\end{align}
The backlash magnitude and the density function are selected as:
\begin{equation} \label{eq:sample}
r_i =
\begin{cases} 
0, & \text{for } i=0 \\ 
r_1+(i-1)\frac{r_n-r_1}{n-1}, & \text{for } i=1,...,n \\ 
\end{cases}
\end{equation}
\begin{equation} \label{eq:sample}
p(r_i)=\lambda e^{-\sigma r_i}
\end{equation}
where $n=30$, $r_1=0.25$, $r_n=7.25$, $\lambda =0.07$, and $\sigma=0.1$. Two magnitude regulators, $\kappa_1=5$ and $\kappa_2=10$, are assigned, and two flag points are set as $v_{f,1}=1.5$ and $v_{f,2}=-0.3$. 

\begin{figure}
\begin{center}
\hspace*{-1.5cm} 
\includegraphics[width= 2.2 in]{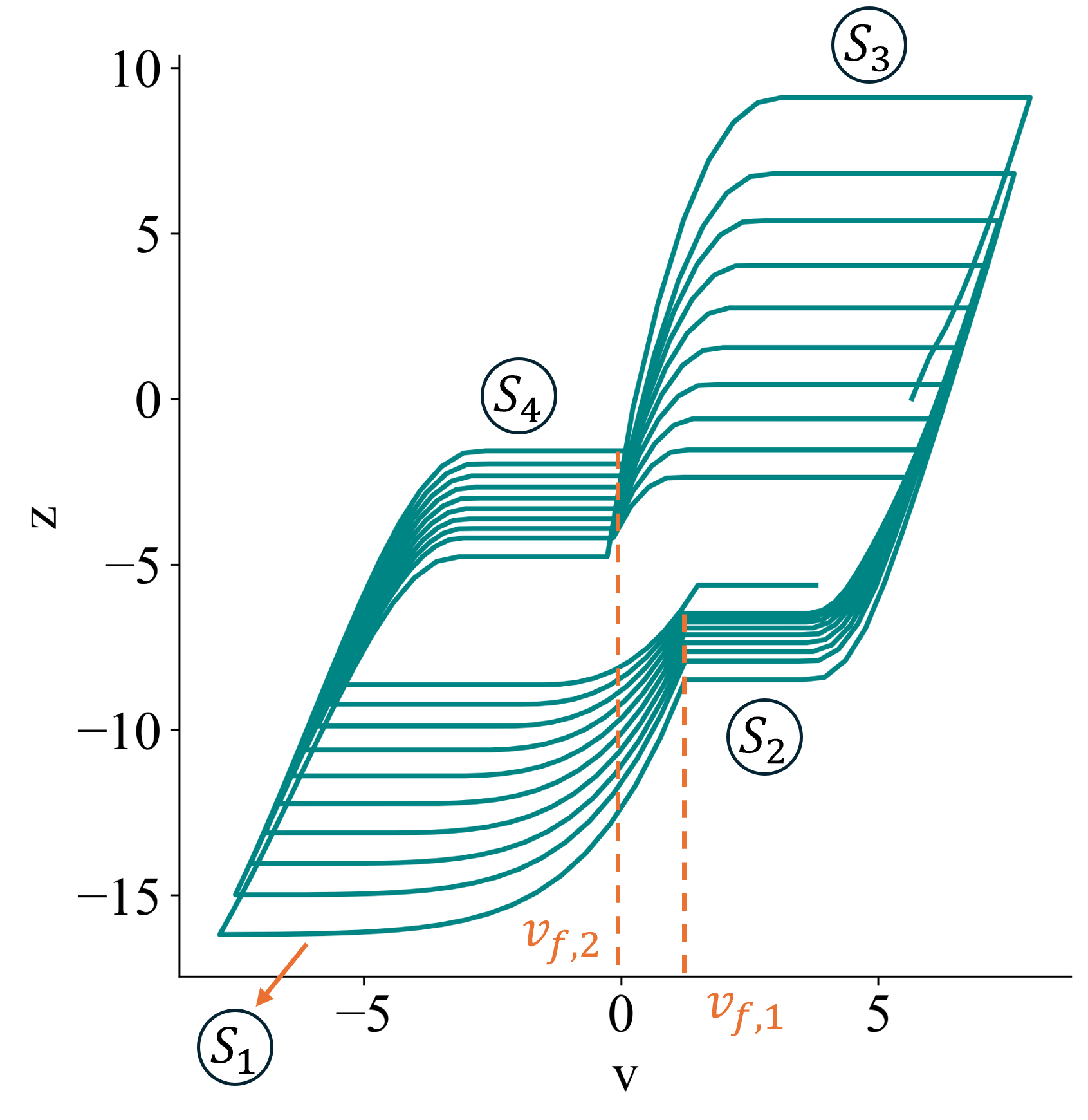}    % The printed column width is 8.4 cm.
\caption{Output of the EGPI model.} 
\label{fig:EGPI_sim}
\end{center}
\end{figure}

The GPI model $z_1(t)$ is shown in Fig.~\ref{fig_m1}(c) and $z_2(t)$ is shown in Fig.~\ref{fig_m1}(d). Since the simulation focuses on illustrating the qualitative input-output relationships, they do not have physical units in Fig.~\ref{fig_m1}(c), (d) and Fig.~\ref{fig:EGPI_sim}. It can be seen that in the increasing and decreasing parts of the inputs of these two GPI models, only one hysteresis dead zone appears. The final output of the EGPI model $z(t)$ that contains the switching mechanism is shown in Fig.~\ref{fig:EGPI_sim}. In the increasing part of the input, there are two stages of hysteresis dead zones $S_1$ and $S_2$; and the same as decreasing part of the input, two stages of hysteresis dead zones are shown in $S_3$ and $S_4$. The hysteresis stage transition points are characterized by two flag points $v_{f,1}$ and $v_{f,2}$. 

The proposed EGPI model has more flexibility in hysteresis modeling over the conventional GPI model. Unlike the GPI model, which enforces continuous Lipschitz continuity throughout the monotonic input domain (\cite{al2010analytical}), the EGPI model enables staged Lipschitz continuity, allowing more accurate representation of segmented hysteresis behaviors observed in I\textsuperscript{2}RIS or any other systems. Moreover, the switching mechanism provides the opportunity to add or remove GPI components and their corresponding flag points in the EGPI model, enabling the modeling of either single-stage or multi-stage hysteresis across the monotonic input domain.

\subsection{Hysteresis Modeling Using EGPI Model}
% To capture the hysteresis behavior of the I\textsuperscript{2}RIS in two orthogonal motion planes, we collected data for yaw and pitch motions in a manner similar to that described in our previous work (\cite{esfandiari2024data}). The input encoder value with two directions to the motors are shown as Fig.~\ref{input_signal}.  In the experimental setup (see Fig.~\ref{fig:exp_setup}), the angular configuration of the distal link—specifically the pitch and yaw bending angles is measured by tracking two ArUco markers attached to the sides of a 3D-printed cube mounted on the I\textsuperscript{2}RIS gripper using two microscopes. A total of four datasets are collected, each corresponding to a different motion direction (positive/negative yaw and pitch).

The motion of the I\textsuperscript{2}RIS during increasing and decreasing input encoder values is shown in Fig.~\ref{fig_m1}(a). Due to material hysteresis during tendon elongation, friction between tendons, friction between tendons and robotic joints, as well as other nonlinear effects, it is clearly identified that two distinct dead zones $S_2$ and $S_3$ appear during the decreasing part of the input (see Fig.~\ref{fig_m1}(a)). To accurately capture the observed hysteresis behavior, the proposed EGPI model is employed. The core idea behind the EGPI modeling approach is to construct an analytical representation of complex, multi-stage hysteresis characteristics. First, because there are two hysteresis stages when the input is decreasing, we design two basic GPI operators $F_{r,1}[v](t)$ and $F_{r,2}[v](t)$ expressed as:
\begin{equation} \label{eq:sample}
F_{r,1}[v](t) =
\begin{cases} 
\text{max}(\gamma_{l} (v(t))-r),w(t_i)), & \text{for } \dot{v}(t)>0 \\ 
\text{min}(\gamma_{r,1} (v(t))+r),w(t_i)), & \text{for } \dot{v}(t)<0 \\ 
w(t_i), & \text{for } \dot{v}(t)=0
\end{cases}
\end{equation}
\begin{equation} \label{eq:sample}
F_{r,2}[v](t) =
\begin{cases} 
\text{max}(\gamma_{l} (v(t)) - r),w(t_i)), & \text{for } \dot{v}(t)>0 \\ 
\text{min}(\gamma_{r,2} (v(t)) + \kappa r),w(t_i)), & \text{for } \dot{v}(t)<0 \\ 
w(t_i), & \text{for } \dot{v}(t)=0
\end{cases}
\end{equation}

where $\gamma_{l}$,$\gamma_{r,1}$ and $\gamma_{r,2}$ are three envelope functions. The function $\gamma_l$ is shared by both $H_{r,1}v$ and $H_{r,2}v$, as there is only one dead zone $S_1$ at the start of the motion (see Fig.~\ref{fig_m1}(a)). In contrast, $\gamma_{r,1}$ and $\gamma_{r,2}$ are assigned to the two operators respectively, in order to capture the two distinct stages of hysteresis $S_2$ and $S_3$ during the decreasing phase. The $\kappa$ is acting as a magnitude regulator (\cite{gao2023modeling}) in order to adjust the dead zone magnitude in $S_3$. By taking the weighted summation of the operators, the two GPI model output are expressed as:
\begin{equation} \label{eq:sample}
y_1(t)=\Pi_1[v](t)=\sum_{i=0}^{n} p(r_i)F_{r,1}[v](t)
\end{equation}
\begin{equation} \label{eq:sample}
y_2(t)=\Pi_2[v](t)=\sum_{i=0}^{n} p(r_i)F_{r,2}[v](t),
\end{equation}
and the EGPI model output is expressed as :
\begin{equation} \label{eq:sample}
y(t) = \Pi[v](t)
     = \begin{cases} 
\Pi_1[v](t), & \text{for } \dot{v}(t)>0\\ 
\Pi_1[v](t), & \text{for } \dot{v}(t)\leq0 \land v(t)> v_f \\ 
\Pi_2[v](t), & \text{for } \dot{v}(t)\leq0 \land v(t)\leq v_f,
\end{cases}
\end{equation}
where the model switches from $\Pi_1[v](t)$ to $\Pi_2[v](t)$ at the flag point $v_f$. The initial $v_f$ is estimated at the first point where $v(\dot{t})$ is approximately zero (\cite{lee2021non}). After constructing the EGPI model, the envelope functions are determined using three linear functions (six parameters) to capture the linearity on both sides of the I\textsuperscript{2}RIS data (see Fig.~\ref{fig_m1}(a)), while reducing computational cost. Envelope functions are expressed as:
\begin{align}
\gamma_{l}(v) &= a_1v + a_2 \\
\gamma_{r,1}(v) &= a_3v + a_4 \\
\gamma_{r,2}(v) &= a_5v + a_6.
\end{align}
The widely adopted density function (\cite{brokate2012hysteresis}) in (15) is parameterized by $\lambda$ and $\sigma$. The backlash values $r_i$ are uniformly sampled from the interval $[r_1, r_n]$ as specified in (14). One magnitude regulator $\kappa$ is set to adjust the backlash in stage $S_3$ (see Fig.~\ref{fig_m1}(a)).
To construct the EGPI model, a total of eleven parameters for each motion direction ($a_1,...,a_6,\lambda, \sigma, r_1, r_n,\kappa$), are determined using the LM method. The objective function is expressed as:
\begin{equation} \label{eq:sample}
L=\sum_{i=0}^{J} (\Pi(v_i)-\theta_j(i))^2,
\end{equation}
where $J$ is the number of the data, and $\theta_j$ is the real measured angle. 

\begin{figure}[!t]
    \centering
    \centerline{\includegraphics[width= 1 \columnwidth]{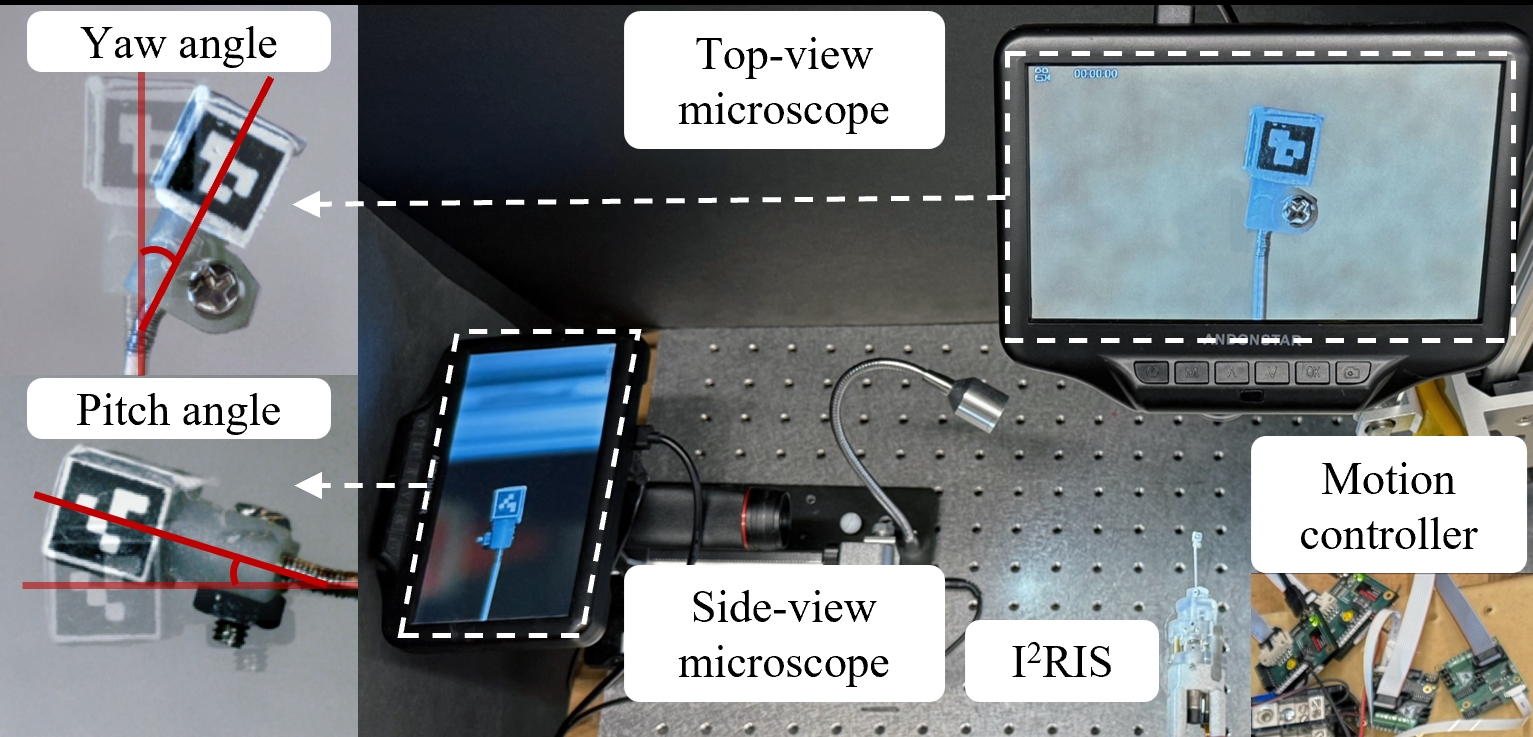}}
    \caption{Experimental setup and bending angle definition.}
    \label{fig:exp_setup}
\end{figure}
\begin{figure}[!t]
\centering
\subfloat[]{\includegraphics[width=1.71in]{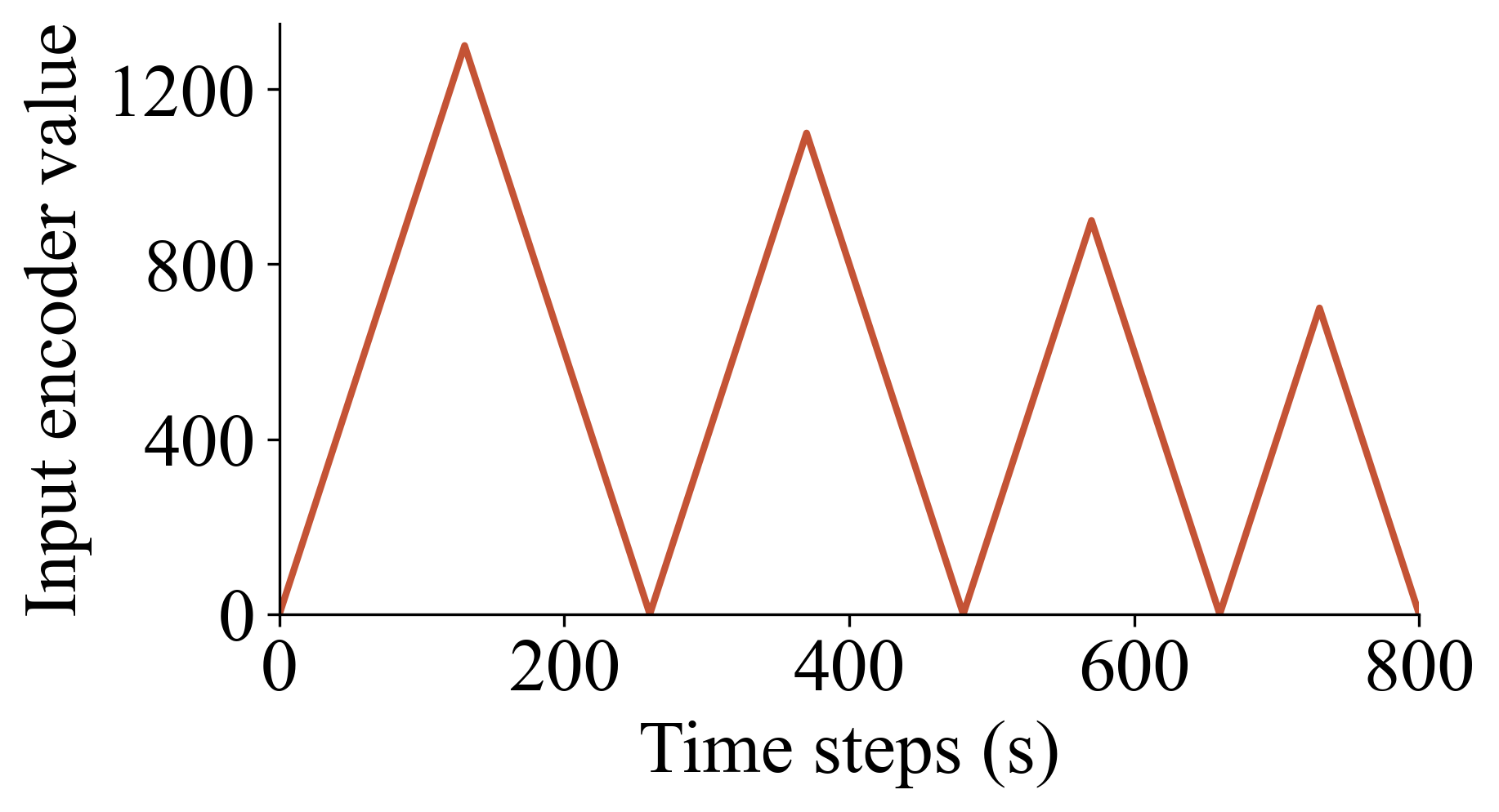}%
\label{input_signal_1}}
\hfil
\subfloat[]{\includegraphics[width=1.71in]{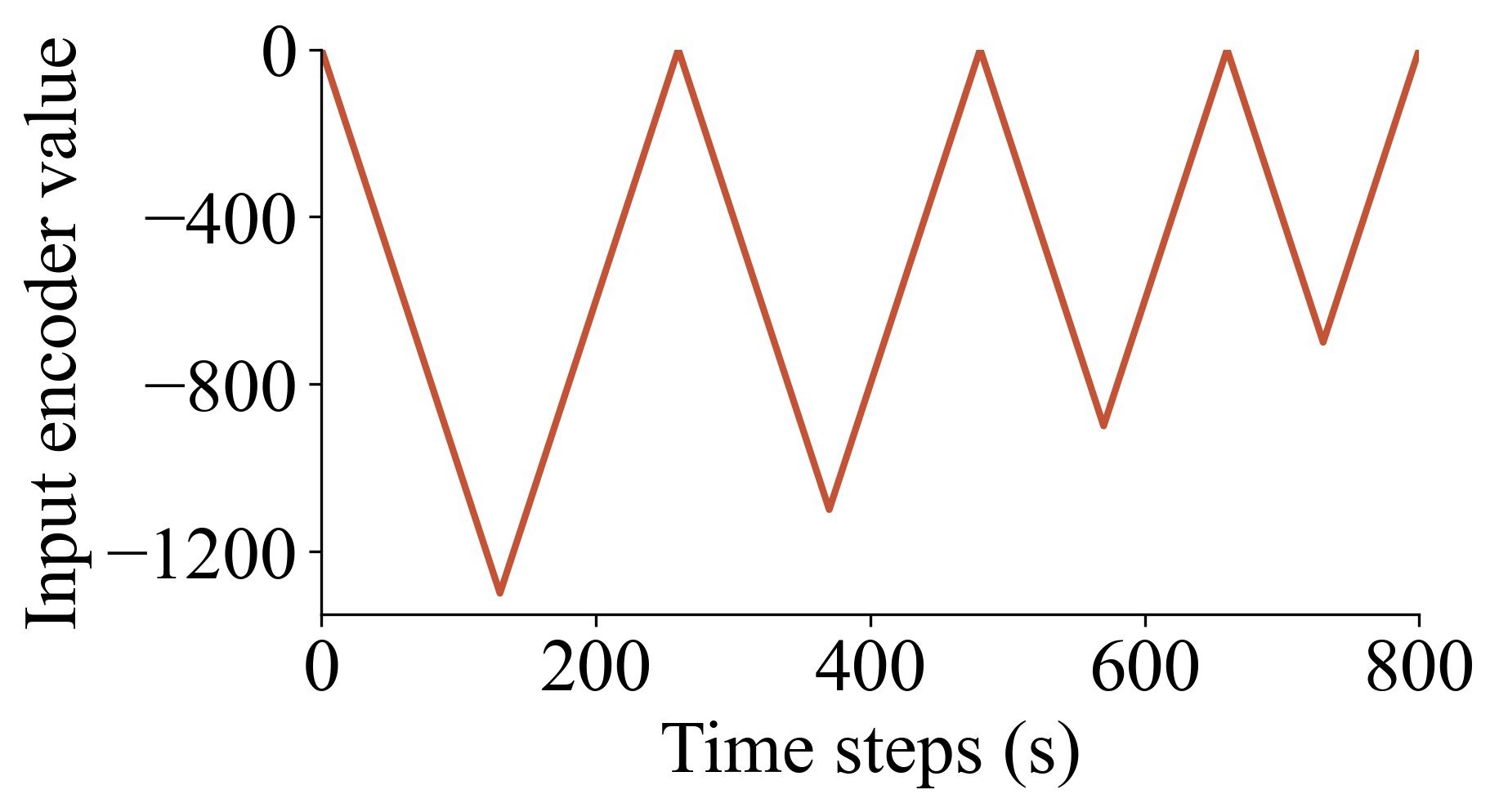}%
\label{input_signal_2}}
\caption{Input encoder value of each motors. (a) Input greater than zero. (b) Input less than zero.}
\label{input_signal}
\end{figure}

\begin{figure*}[!t]
\centering
\subfloat[]{\includegraphics[width=1.75in]{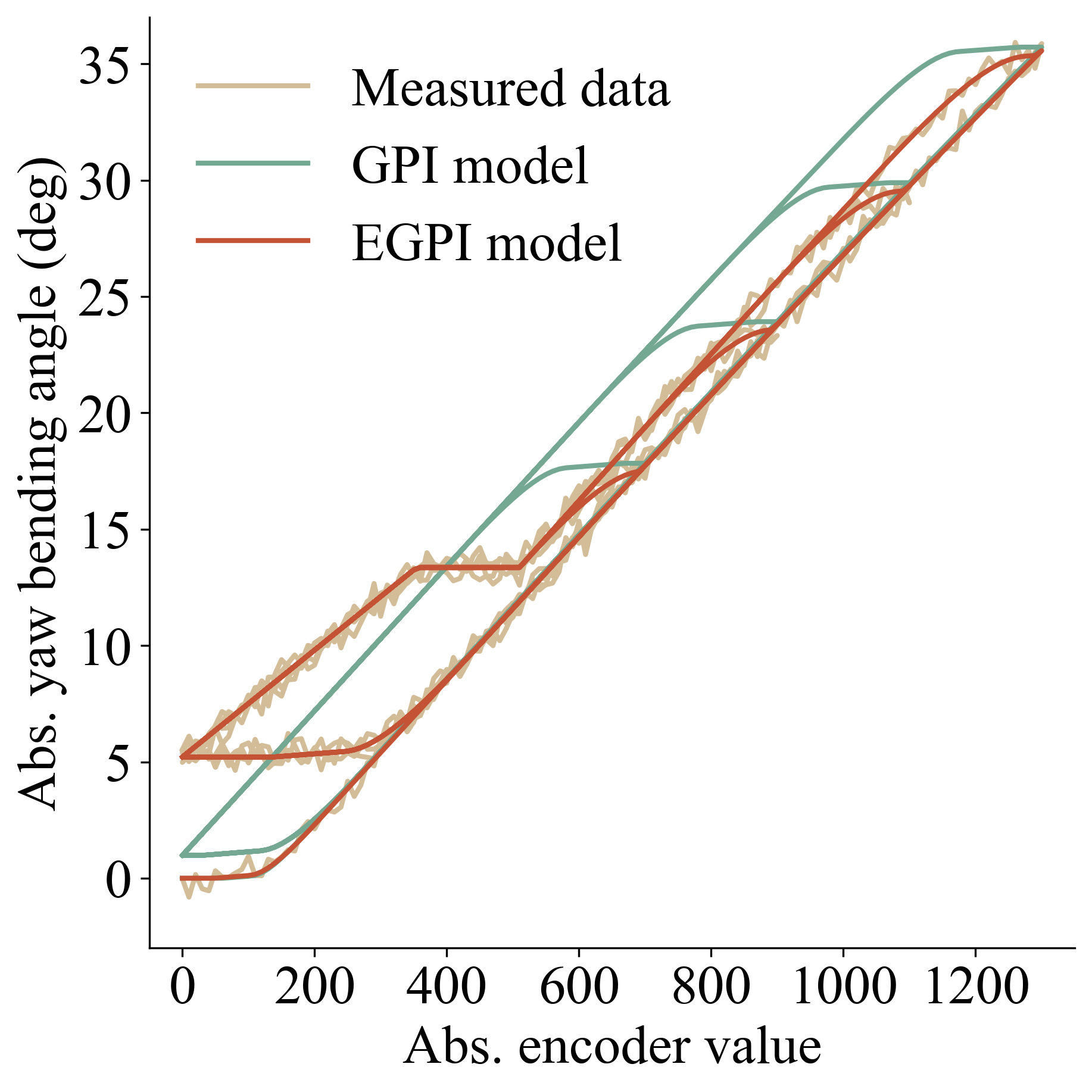}%
\label{result_a}}
\hfil
\subfloat[]{\includegraphics[width=1.75in]{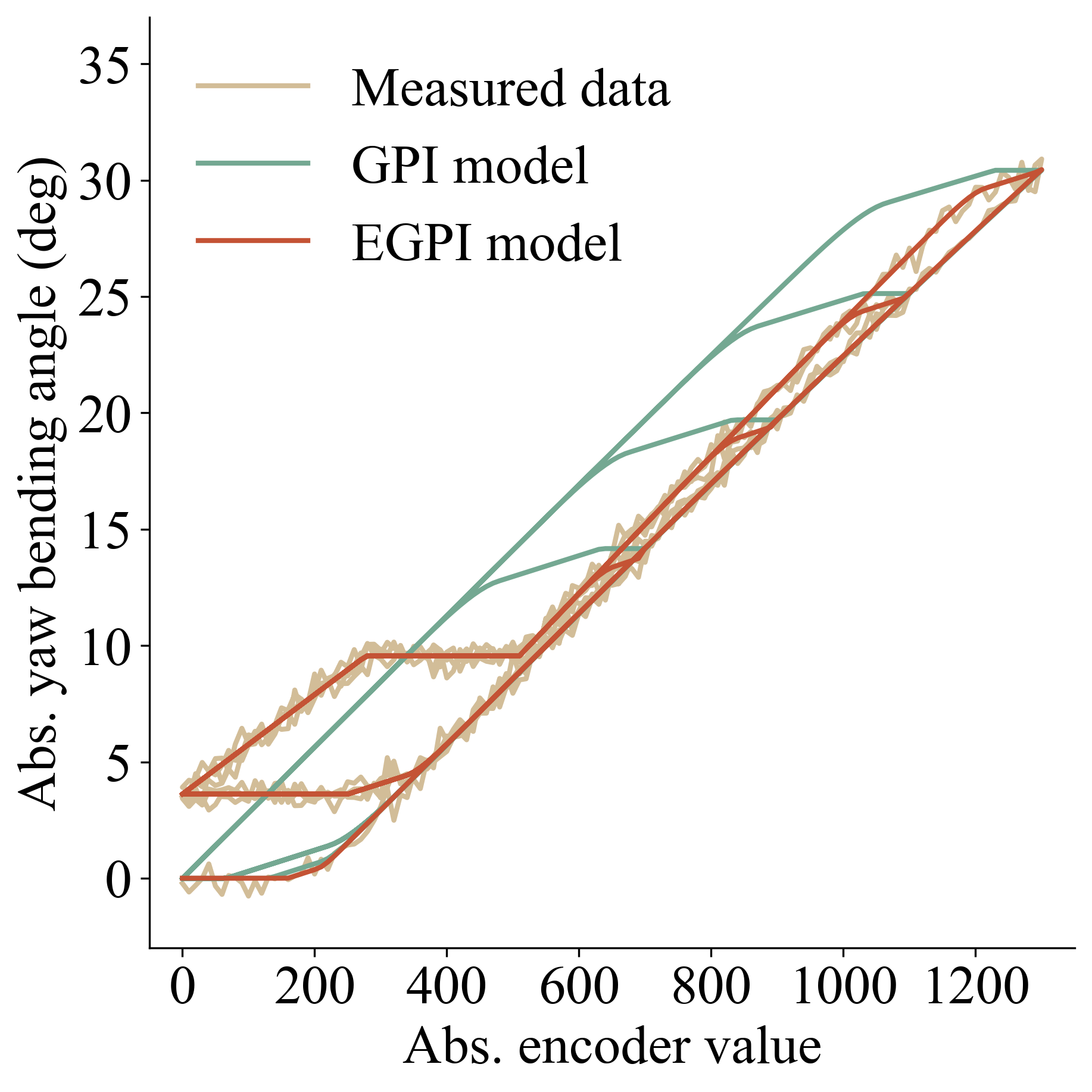}%
\label{result_b}}
\hfil
\subfloat[]{\includegraphics[width=1.75in]{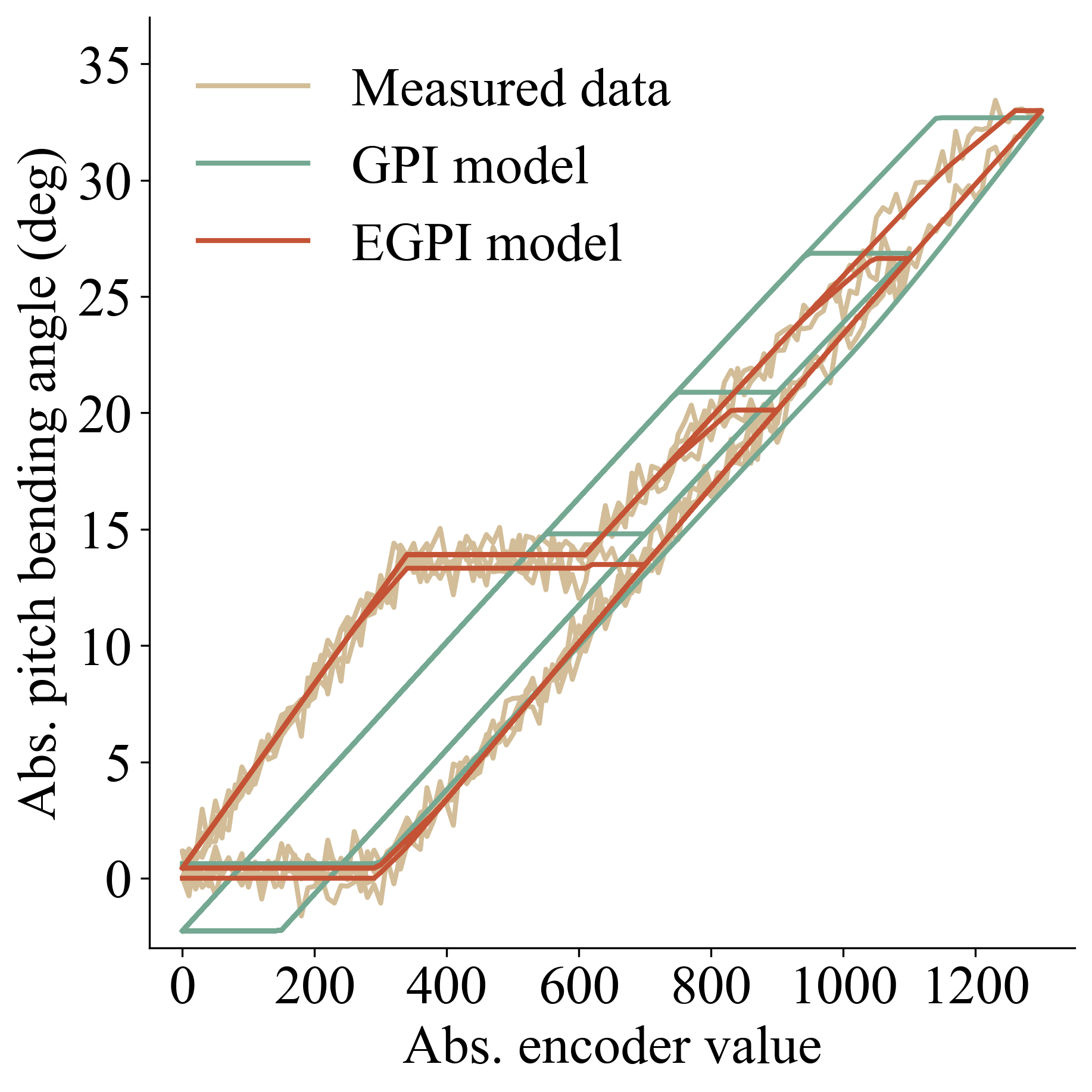}%
\label{result_c}}
\hfil
\subfloat[]{\includegraphics[width=1.75in]{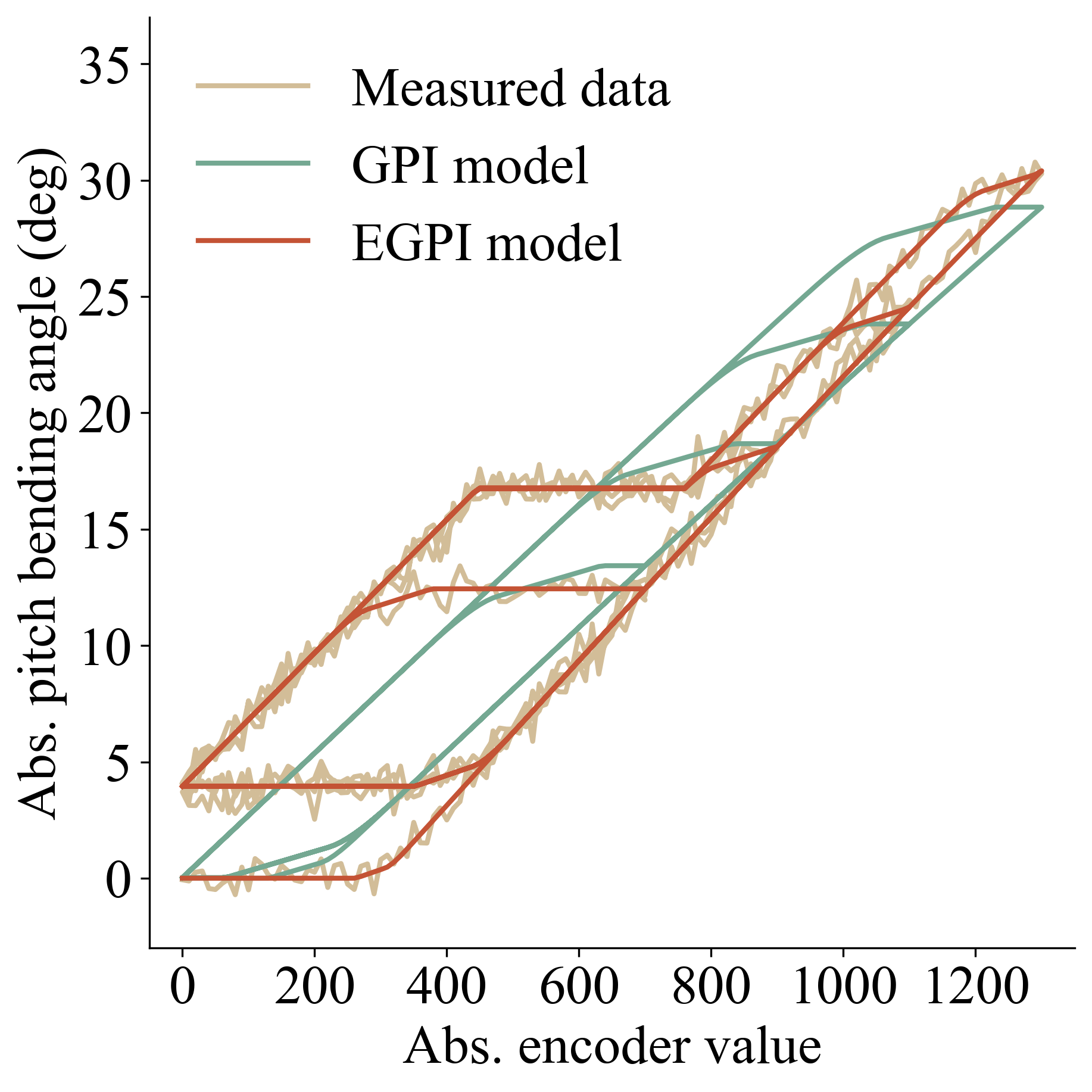}%
\label{result_d}}

\subfloat[]{\includegraphics[width=1.76in]{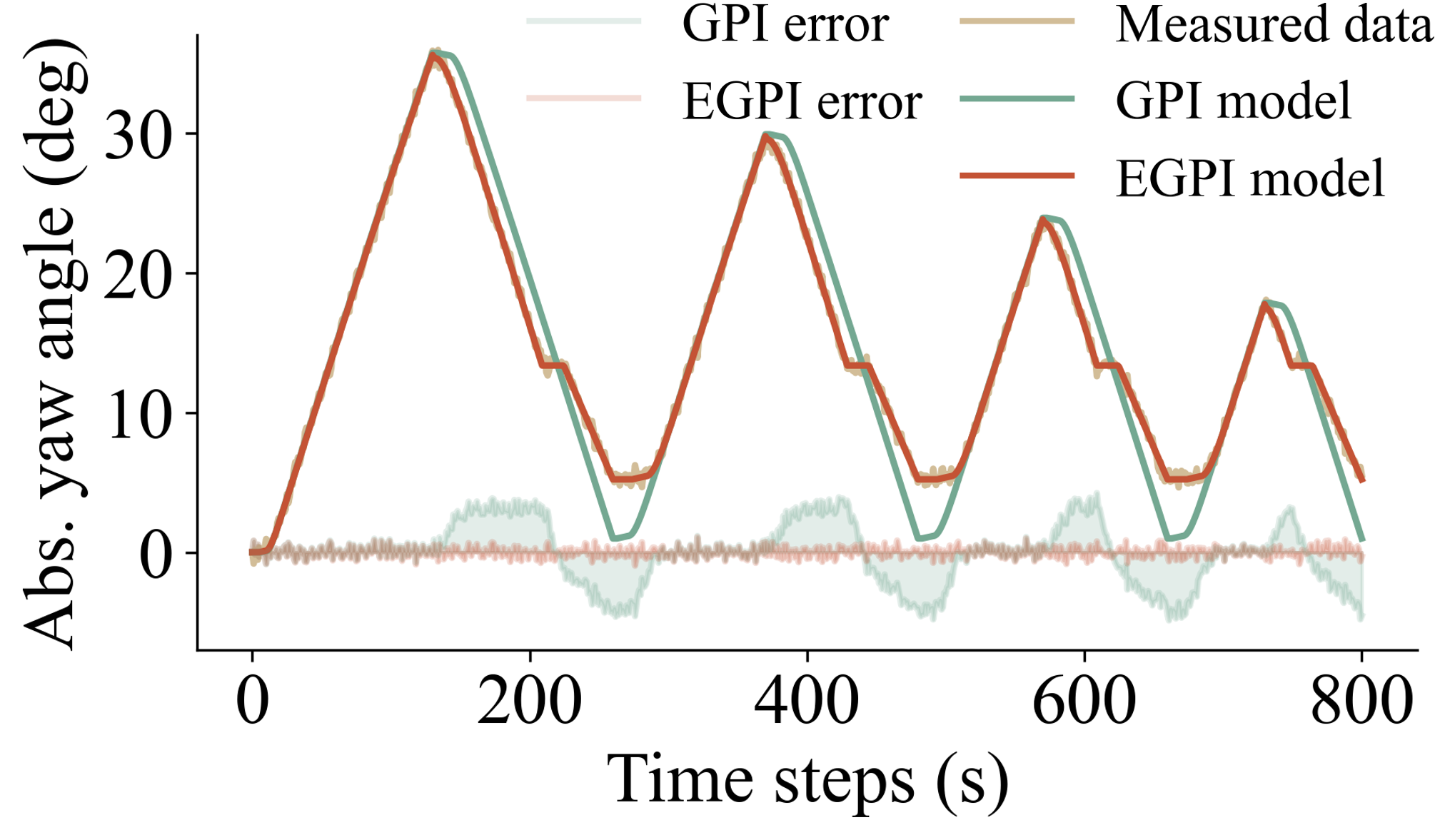}%
\label{result_e}}
\hfil
\subfloat[]{\includegraphics[width=1.76in]{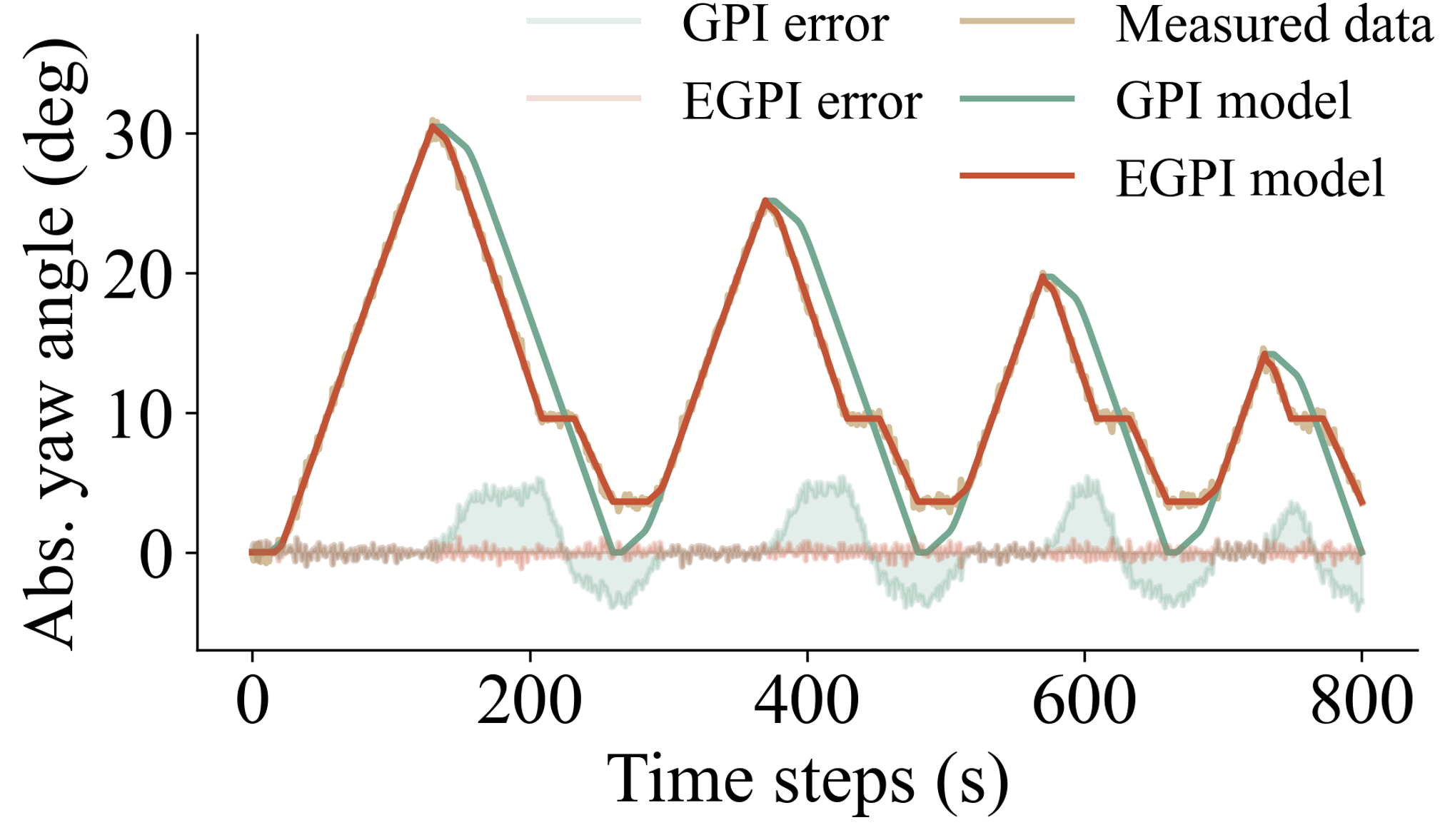}%
\label{result_f}}
\hfil
\subfloat[]{\includegraphics[width=1.76in]{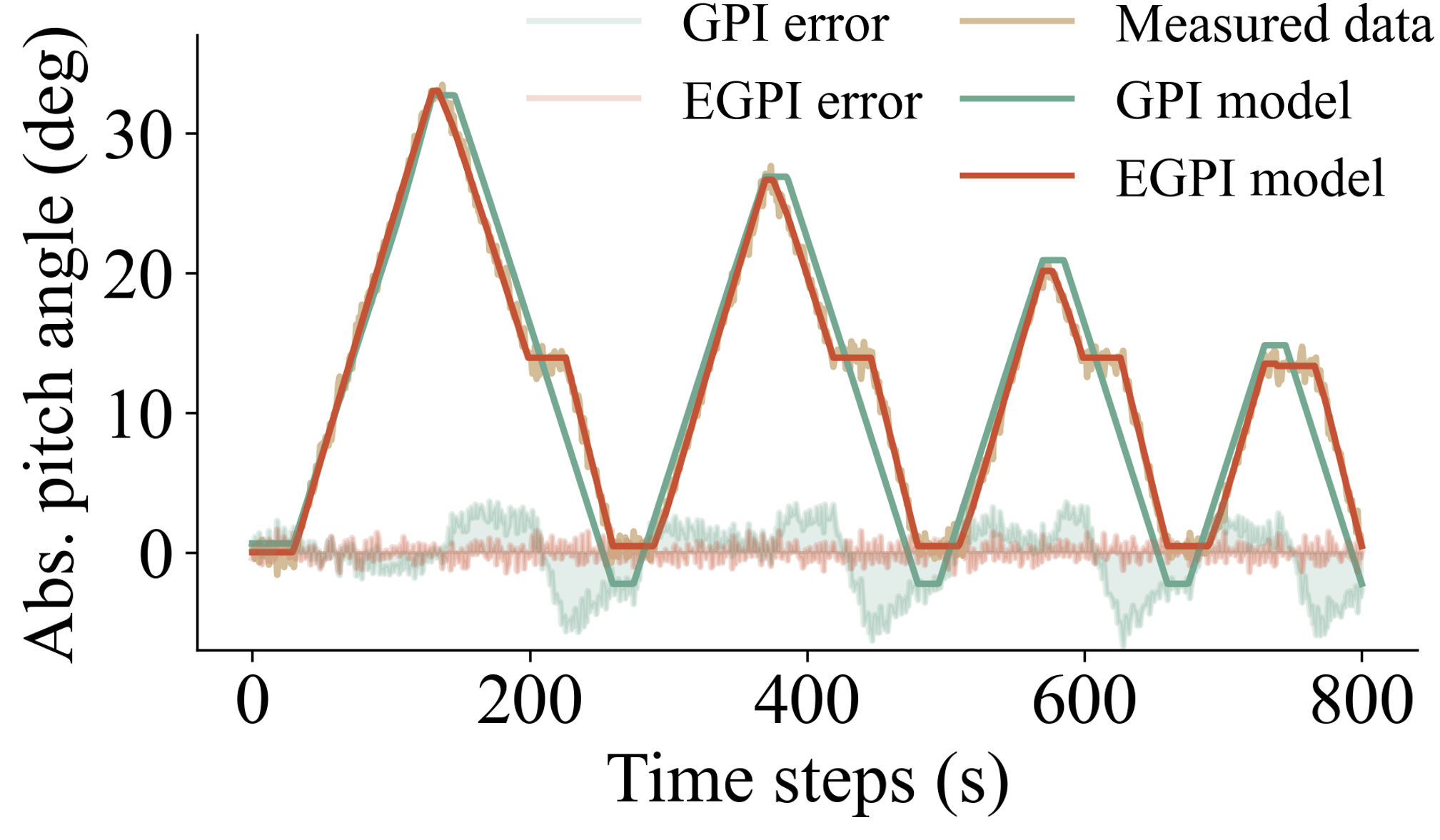}%
\label{result_g}}
\hfil
\subfloat[]{\includegraphics[width=1.76in]{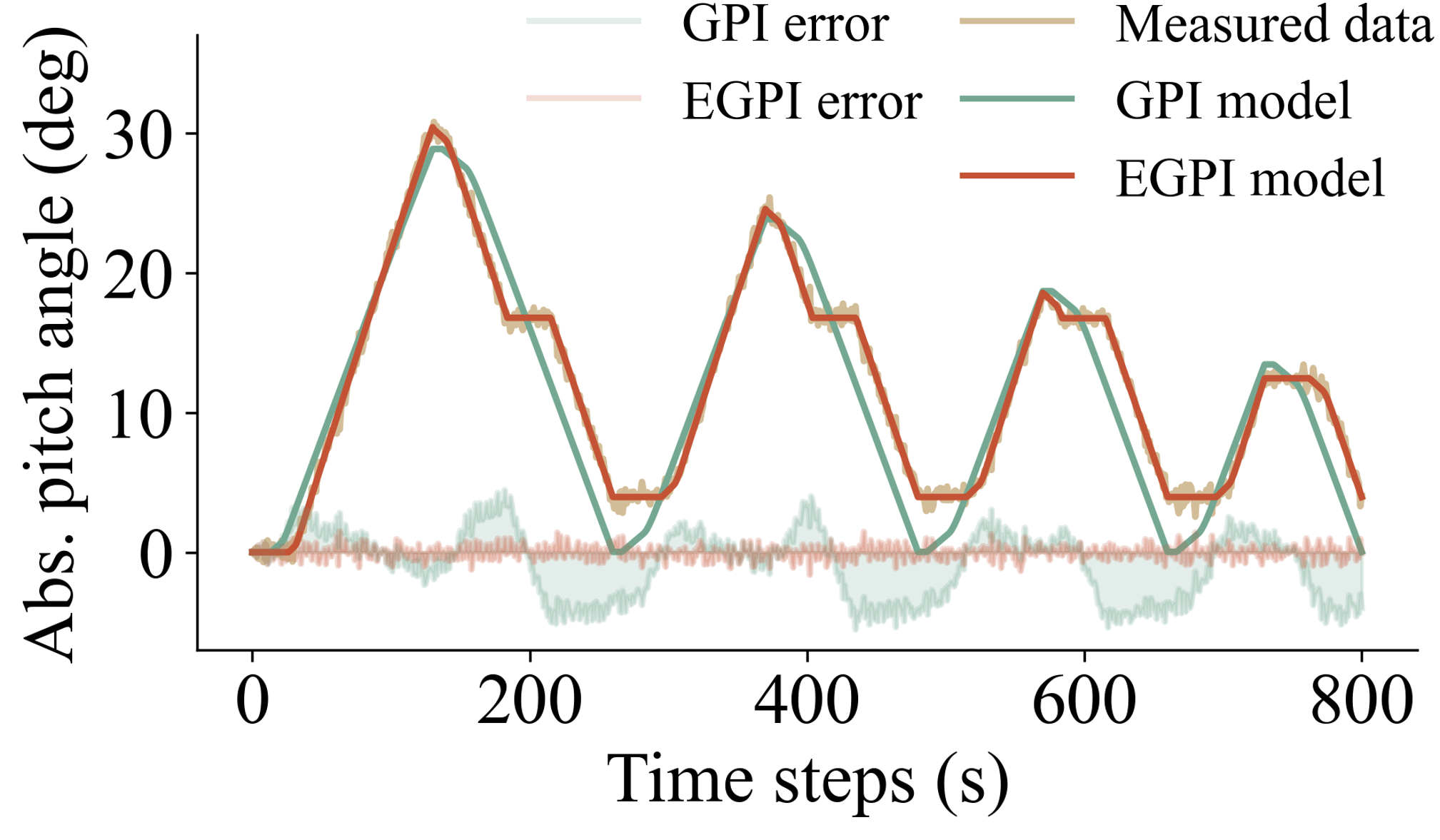}%
\label{result_h}}

\caption{Hysteresis modeling results. (a) Yaw motion (input$>$0): yaw angle vs. input. (b) Yaw motion (input$<$0): yaw angle vs. input. (c) Pitch motion (input$>$0): pitch angle vs. input. (d) Pitch motion (input$<$0): pitch angle vs. input. (e) Yaw motion (input$>$0): yaw angle vs. time. (f) Yaw motion (input$<$0): yaw angle vs. time. (g) Pitch motion (input$>$0): pitch angle vs. time. (h) Pitch motion (input$<$0): pitch angle vs. time.}
\label{result}
\end{figure*}

\section{Experimental Results}
To capture the hysteresis behavior of the I\textsuperscript{2}RIS, we developed an experimental setup (see Fig.~\ref{fig:exp_setup}) to collect data for two orthogonal motion planes: yaw and pitch in a manner similar to that described in our previous work (\cite{esfandiari2024data}).   The actuation unit comprises two DC servo motors (DCX8M, Maxon Motor Inc.) with an integrated reduction gear (GPX8, gear ratio 64:1), encoders (ENX 8MAG 256 pulses/revolution), and controllers (EPOS2 24/2). The low-level motor position control and the high-level command generation are implemented in C++ and Python, respectively. Communications are handled in ROS1 Noetic. In the experimental setup, the pitch and yaw bending angles of the dexterous distal unit are measured by tracking two ArUco markers attached to the sides of a 3D-printed cube mounted on the gripper using two microscopes. 

A total of four datasets were collected, each corresponding to a different motion direction (positive/negative yaw and pitch). The input encoder signals for both directions of a single motor are shown in Fig.~\ref{input_signal}.

The resulting EGPI hysteresis model outputs are presented in Fig.~\ref{result}, along with a comparison with the GPI model results. To focus on the input and output relationships, we correlate the absolute encoder value with the absolute bending angle for better visualization. For example, in Fig.~\ref{result}(b), the actual motion corresponds to the yaw motor movement with an input encoder value less than zero, as shown in Fig.~\ref{input_signal}(b). In this case, we use the absolute value of the encoder value to visualize the relationship with the positive yaw bending angle.
To quantitatively evaluate the models performance, three metrics are defined as follows: Root Mean Square Error (RMSE) 
\begin{align}
RMSE = \sqrt{\frac{1}{N} \sum_{i=1}^{N} (\theta_i - \hat{\theta}_i)^2}, i=1,2,...,N,
\end{align}
Normalized Root Mean Square Error (NRMSE)
\begin{align}
NRMSE = \frac{RMSE}{\theta_{max}-\theta_{min}}\times 100\%,
\end{align}
Maximum Absolute Error (MAE)
\begin{align}
MAE = \text{max}\left| \theta_i - \hat{\theta}_i \right|,i=1,2,...,N,
\end{align}
where $\theta_i$ is the I\textsuperscript{2}RIS data, $\hat{\theta}_i$ is the modeling result and $N$ is the number of data points. 

The quantitative performance comparison between the EGPI and GPI models is presented in Table~\ref{tab:err_yaw} and Table~\ref{tab:err_pitch} for the motion of two distinct motors. Overall, the EGPI model outperforms the GPI model in terms of accuracy and fitting quality, as clearly demonstrated in Fig.~\ref{result}. The orange line, representing the EGPI model, shows a closer fit to the measured data compared to the green line for the GPI model. This trend is consistently observed across all four motion directions. It is evident from Fig.~\ref{result}(e)--(h) that the GPI model exhibits more prominent error fluctuations compared to the EGPI model.

Moreover, the EGPI model consistently yields lower RMSE, NRMSE, and MAE values compared to the GPI model. For example, in the yaw motor motion with positive input, the RMSE for EGPI is $0.36^\circ$, significantly lower than the GPI model's RMSE of $2.22^\circ$. This reduction in error is particularly evident in the decreasing regions of the input, as shown in Fig.~\ref{result}(a), where the EGPI model captures the hysteresis behavior more accurately. The main reason for this improvement is that the EGPI model can switch its hysteresis representation across different dead-zone stages, a capability that the GPI model lacks.

Furthermore, when considering the physical meaning of the data, the EGPI model provides a more reliable representation of the I\textsuperscript{2}RIS hysteresis across all motion scenarios. These results highlight the effectiveness of the EGPI model in capturing complex hysteresis behavior, making it a valuable method for improving accuracy in modeling similar characteristics.

\renewcommand\cellalign{cc}
\renewcommand\cellgape{\Gape[1pt]}
\renewcommand{\arraystretch}{1.0}
\begin{table}[h!]
\centering
\caption{Quantitative performance of EGPI and GPI model for yaw motor motion}
\label{tab:err_yaw}
\setlength{\tabcolsep}{3pt}  % 控制列之间的间距，默认是6pt，可以减小
% \begin{tabular}{@{}c ccc ccc ccc ccc@{}}
\begin{tabular}{
    @{}c
    @{\hskip 4pt}ccc
    @{\hskip 4pt}ccc@{}
}
\toprule
\multirow{2}{*}{\makecell[c]{\textbf{Modeling} \\ \textbf{Method}}} &
\multicolumn{3}{c}{\textbf{Positive input}} &  
\multicolumn{3}{c}{\textbf{Negative input}} \\
& \textbf{RMSE} & \textbf{NRMSE} & \textbf{MAE}  
& \textbf{RMSE} & \textbf{NRMSE} & \textbf{MAE} \\
\midrule
\textbf{EGPI} & \textbf{0.36$^\circ$} & \textbf{0.98\%} & \textbf{1.07$^\circ$} & \textbf{0.35$^\circ$} & \textbf{1.12\%} & \textbf{1.15$^\circ$} \\
\textbf{GPI}  & 2.23$^\circ$ & 6.07\% & 4.88$^\circ$ & 2.38$^\circ$ & 7.50\% & 5.37$^\circ$ \\
\bottomrule
\end{tabular}
\end{table}

\begin{table}[h!]
\centering
\caption{Quantitative performance of EGPI and GPI model for pitch motor motion}
\label{tab:err_pitch}
\setlength{\tabcolsep}{3pt}  % 控制列之间的间距，默认是6pt，可以减小
% \begin{tabular}{@{}c ccc ccc ccc ccc@{}}
\begin{tabular}{
    @{}c
    @{\hskip 4pt}ccc
    @{\hskip 4pt}ccc@{}
}
\toprule
\multirow{2}{*}{\makecell[c]{\textbf{Modeling} \\ \textbf{Method}}} &
\multicolumn{3}{c}{\textbf{Positive input}} &  
\multicolumn{3}{c}{\textbf{Negative input}} \\
& \textbf{RMSE} & \textbf{NRMSE} & \textbf{MAE}  
& \textbf{RMSE} & \textbf{NRMSE} & \textbf{MAE} \\
\midrule
\textbf{EGPI} & \textbf{0.58$^\circ$} & \textbf{1.65\%} & \textbf{1.62$^\circ$} & \textbf{0.49$^\circ$} & \textbf{1.12\%} & \textbf{1.62$^\circ$} \\
\textbf{GPI}  & 2.51$^\circ$ & 7.16\% & 6.73$^\circ$ & 2.64$^\circ$ & 8.38\% & 5.53$^\circ$ \\
\bottomrule
\end{tabular}
\end{table}

\section{Conclusion and future work}

In this work, we proposed an EGPI model to effectively capture the hysteresis behavior of I\textsuperscript{2}RIS. Simulation results demonstrated its capability to model multi-stage hysteresis through a novel switching mechanism. Experiments on I\textsuperscript{2}RIS data further validated the model, showing significantly better fitting accuracy than the GPI model across three error metrics. These results highlight the strong potential of the EGPI model to represent more complex hysteresis behaviors in other robotic systems. In future work, we plan to enhance the EGPI model by introducing mathematical constraints to ensure continuity across switching points between submodels. Next, we will develop an analytical inverse of the EGPI model to establish a mapping from the I\textsuperscript{2}RIS joint space to the actuation space. With the inverse model, a compensation strategy can be explored to enable precise control of I\textsuperscript{2}RIS motion toward the desired joint angles.

%%%%%%%%%%%%%%%%%%%%%%%%%%%%%%%%%%%%%%%%%%%%%%%%%%%%%%%%%%%%%%%%%%%%%%%%%
%%%%%%%%%%%%%%%%%%%%%%%%%%%%%%%%%%% THE END %%%%%%%%%%%%%%%%%%%%%%%%
%%%%%%%%%%%%%%%%%%%%%%%%%%%%%%%%%%%%%%%%%%%%%%%%%%%%%%%%%%%%%%%%%%%%%%%%%
\bibliography{ifacconf}             % bib file to produce the bibliography
                                                                                             
\end{document}